\documentclass[a4paper,conference]{IEEEtran}
\IEEEoverridecommandlockouts
\usepackage{afterpage}

\usepackage{cite}
\usepackage{amsmath,amssymb,amsfonts}
\usepackage{algorithmic}
\usepackage{graphicx}
\usepackage{textcomp}
\usepackage{xcolor}
\usepackage{multirow}
\usepackage{subfig}
\usepackage[normalem]{ulem} 
\usepackage{balance}
\usepackage{hyperref}
\usepackage{placeins}

\def\BibTeX{{\rm B\kern-.05em{\sc i\kern-.025em b}\kern-.08em
    T\kern-.1667em\lower.7ex\hbox{E}\kern-.125emX}}
    
\begin{document}

\title{Person Detection and Tracking from an Overhead Crane LiDAR \\

\thanks{This work was funded in part by Business Finland under the TwinFlow project (7374/31/2023), and in part by the Finnish Centre for Economic Development, Transport and the Environment under the TECHBOOST project (0124/05.02.09/2023A).}
}

\author{\IEEEauthorblockN{1\textsuperscript{st} Nilusha Jayawickrama}
\IEEEauthorblockA{\textit{School of Engineering} \\
\textit{Aalto University}\\
Espoo, Finland \\
nilusha.jayawickrama@aalto.fi}
\and
\IEEEauthorblockN{2\textsuperscript{nd} Henrik Toikka}
\IEEEauthorblockA{\textit{School of Electrical Engineering} \\
\textit{Aalto University}\\
Espoo, Finland \\
henrik.toikka@aalto.fi}
\and
\IEEEauthorblockN{3\textsuperscript{rd} Risto Ojala}
\IEEEauthorblockA{\textit{School of Engineering} \\
\textit{Aalto University}\\
Espoo, Finland \\
risto.j.ojala@aalto.fi}
}

\maketitle

\begin{abstract}
This paper investigates person detection and tracking in an industrial indoor workspace using a LiDAR mounted on an overhead crane. The overhead viewpoint introduces a strong domain shift from common vehicle-centric LiDAR benchmarks, and limited availability of suitable public training data. Henceforth, we curate a site-specific overhead LiDAR dataset with 3D human bounding-box annotations and adapt selected candidate 3D detectors under a unified training and evaluation protocol. We further integrate lightweight tracking-by-detection using AB3DMOT and SimpleTrack to maintain person identities over time. Detection performance is reported with distance-sliced evaluation to quantify the practical operating envelope of the sensing setup. The best adapted detector configurations achieve average precision (AP) up to 0.84 within a 5.0 m horizontal radius, increasing to 0.97 at 1.0 m, with VoxelNeXt and SECOND emerging as the most reliable backbones across this range. The acquired results contribute in bridging the domain gap between standard driving datasets and overhead sensing for person detection and tracking. We also report latency measurements, highlighting practical real-time feasibility. Finally, we release our dataset and implementations in \href{https://github.com/nilushacj/O-LiPeDeT-Overhead-LiDAR-Person-Detection-and-Tracking}{GitHub} to support further research.
\end{abstract}


\section{Introduction}\label{sec:intro}

With the increasing adoption of automation in industrial environments, ensuring the safety of human workers operating in close proximity to such systems has become a crucial challenge. In industrial settings such as factories and warehouses, failures to reliably detect nearby personnel can lead to severe injuries, equipment damage, and costly operational downtime. A suitable option for mitigating these risks is to integrate the ability to accurately detect and localize people within the workspace, enabling downstream components such as monitoring systems or collision avoidance modules to react appropriately. Reliable person detection thus forms the basis for situational awareness and safe decision-making in automated and semi-autonomous industrial operations.

Various approaches have been studied in prior works. Models for pedestrian detection in autonomous driving have been commonly explored \cite{Fang2022occlusion}, while proof of fine-tuning these models for other settings such as indoor environments are also evident \cite{Linder2021cross}\cite{ahmed2021top}. Camera-based detection has been a popular choice, but recently LiDAR-only 3D detection (from point clouds) have been widely adopted. Specifically, an overhead-view LiDAR (OLiDAR) is an attractive option for human-aware safety and automation in industrial spaces where both people and machinery operate in close proximity. Some key benefits of a LiDAR in this context are that it is less sensitive to lighting, provides direct 3D geometry, and is naturally more privacy-preserving as it does not capture identity (ID) details in the same way as video feed. However, some non-trivial challenges of a LiDAR include sparse points on small targets like people, and point density variation with range and incidence angle. Additionally, there is also a lack of publicly available overhead-view LiDAR datasets that can be readily utilized for detection or tracking implementations.

\begin{figure*}[!ht]
    \centering

    \subfloat[PointPillars\label{pp_small}]{
        \includegraphics[width=0.186\textwidth]{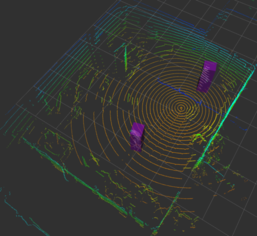}
    }
    \subfloat[SECOND\label{second_small}]{
        \includegraphics[width=0.186\textwidth]{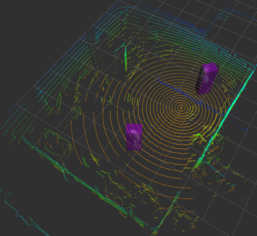}
    }
    \subfloat[PV-RCNN\label{pvrcnn_small}]{
        \includegraphics[width=0.186\textwidth]{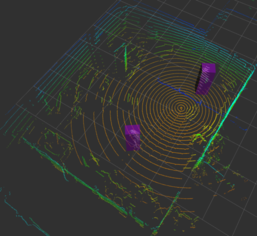}
    }
    \subfloat[VoxelNeXt\label{voxelnext_small}]{
        \includegraphics[width=0.186\textwidth]{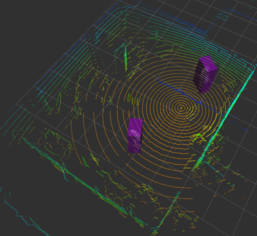}
    }
    \subfloat[Voxel RCNN\label{voxelrcnn_small}]{
        \includegraphics[width=0.186\textwidth]{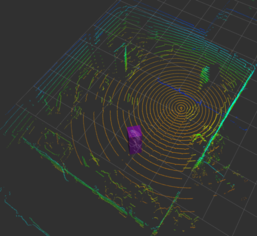}
    }

    \vspace{0.1mm}

    \subfloat[PointPillars\label{pp_large}]{
        \includegraphics[width=0.186\textwidth]{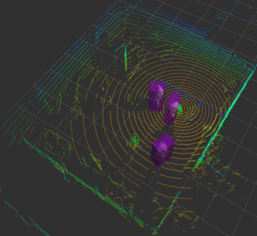}
    }
    \subfloat[SECOND\label{second_large}]{
        \includegraphics[width=0.186\textwidth]{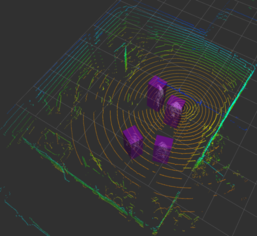}
    }
    \subfloat[PV-RCNN\label{pvrcnn_large}]{
        \includegraphics[width=0.186\textwidth]{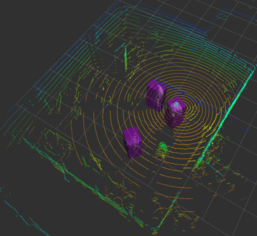}
    }
    \subfloat[VoxelNeXt\label{voxelnext_large}]{
        \includegraphics[width=0.186\textwidth]{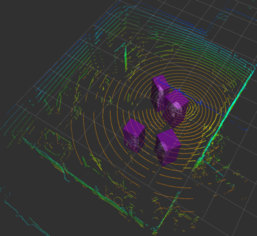}
    }
    \subfloat[Voxel RCNN\label{voxelrcnn_large}]{
        \includegraphics[width=0.186\textwidth]{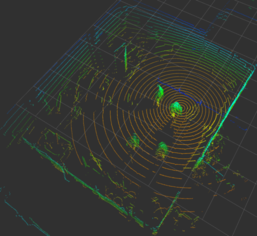}
    }

    \caption{Qualitative predictions from the adapted detection modules for two sample frames, comprising two (\textbf{(a)}-\textbf{(e)}) and four (\textbf{(f)}-\textbf{(j)}) human targets respectively.}

    \label{sample_preds}
\end{figure*}

This paper introduces an overhead-view LiDAR person detection benchmark for indoor workspace monitoring through an adaptation and evaluation of suitable 3D LiDAR detectors on a site-specific self-annotated dataset. Specifically, we retrain multiple machine learning models for the overhead viewpoint and compare their accuracy and runtime characteristics under a consistent evaluation protocol. Sample results are presented in Figure \ref{sample_preds}. Moreover, we analyze performance as a function of target distance to indicate the practical operating envelope of the sensing setup which we installed in an overhead crane environment.

Henceforth, the \textbf{main contributions} for this paper include:
\textbf{(i)} transfer learning and evaluation of relevant LiDAR models for person detection based on an overhead-view LiDAR viewpoint within an industrial crane workspace, \textbf{(ii)} a site-specific 3D annotated point cloud dataset of human targets, and \textbf{(iii)} distance-sliced evaluation to demonstrate detection feasibility and potential operating envelope.

\section{Related work}\label{sec:sota}

\subsection{LiDAR-based detection}\label{subsec:sotadetection}

Various approaches have been investigated for reliable detection in 3D space based on LiDAR input. The earliest explored category is projection-based methods, in which the 3D point cloud is projected to a 2D representation and a 2D CNN is applied for object detection. 
Most methodologies are typically front-view (FV) projection-based with representative models including FVNet \cite{zhou2019fvnet} and LaserNet \cite{meyer2019lasernet}. The main drawbacks of these techniques are the high scale variation introduced by perspective projection, projection induced distortion of local neighbourhoods (as 2D pixel adjacency does not always correspond to 3D point adjacency), and high sensitivity to occlusions, which hinder accurate localization, particularly for small or distant objects. As a result, projection-based methods were gradually superseded by voxel-based and point-based methods. 

Voxel-based methods focus on structuring the 3D space into voxel grid representations and applying 3D convolution (CNN) to enable detection. VoxelNet \cite{zhou2018voxelnet}, commonly regarded as the most influential voxel-based approach, introduced a framework which combines voxel feature encoding of the raw point clouds with 3D CNN for object detection. Despite its effectiveness, VoxelNet is computationally heavy and therefore, subsequently developed models such as SECOND (SEC) \cite{yan2018second} replaced dense convolutions with sparse convolutions. To further enhance both accuracy and efficiency, new voxel-based architectures continued to be introduced: \textbf{(i)} Voxel RCNN \cite{deng2021voxelrcnn} applied region of interest (RoI) pooling to refine feature extraction, \textbf{(ii)} CIA-SSD \cite{zheng2021cia} combined a lightweight feature learning network with confidence refining on the bird's-eye view (BEV) grid to improve detection reliability, and \textbf{(iii)} VoxelNeXt \cite{chen2023voxelnext} delivered a fully sparse pipeline that completely avoided dense conversions. Furthermore, the work in \cite{Lang2019CVPR} introduced PointPillars which discretizes the point cloud into vertical pillars instead of full 3D voxels. This results in a 2D pseudo-image where each pillar is encoded with a lightweight learned feature extractor that allows the model to solely use 2D CNN for detection.  

Point-based methods aim to directly operate on raw point clouds for point-level feature extraction without requiring any projection or voxelization, and rely on modeling local and global geometric relationships. Popular representative models include: \textbf{(i)} PointRCNN \cite{shi2019pointrcnn} which employs bottom-up proposal generation from raw point clouds, \textbf{(ii)} Point-GNN which encodes the raw point cloud as a graph and uses graph neural networks for bounding box prediction \cite{shi2020pointgnn}, and \textbf{(iii)} Pointformer \cite{pan2021pointformer} which learns features from raw points using a transformer, yielding a richer context and capturing long-range dependencies. Despite high geometric precision, a key limitation with point-based methods is that they typically require comparatively heavier processing as a result of the irregular structure of point clouds and thus more complex neighbourhood queries.

Most detectors which studied performance of human targets have been in the contexts of vulnerable road user detection and surveillance applications \cite{wang2022lidar}\cite{blanch2024surveillance}. Early methods addressed person detection by leveraging geometric features and point cloud clustering, and typically faced challenges under dense crowds. More recent models aimed at better localizing and maintaining the IDs of human targets in dynamic scenes, in which they emphasized the trade-offs between accuracy and speed to balance the real-time performance with adequate capability to capture the geometric structure of humans \cite{na2023real}. 

\subsection{Multi-object tracking models}\label{subsec:sotatrack}

Multi-object tracking (MOT) models estimate the trajectories and IDs of multiple objects over time, commonly by associating detections across consecutive frames \cite{zhang2022bytetrack}.

Most classical tracking approaches rely on motion filtering (e.g. Kalman filtering), combined with data association strategies such as Hungarian matching, nearest neighbour or distance-based matching, and IoU gating \cite{luo2021multiple}. AB3DMOT \cite{weng2020ab3dmot} is a widely adopted model that works by tracking 3D bounding boxes using a Kalman filter (KF) motion model and IoU-based association. Similarly, SimpleTrack \cite{pang2022simpletrack} relies on classical MOT design choices within the same tracking-by-detection family, showing that careful selection of motion and association components can yield a simple yet effective baseline. Despite being early techniques in the field, these models remain widely used in real-time systems as a result of being lightweight and computationally efficient. The key drawback however is that there is no learning from temporal context, and therefore the approach is unable to recognize patterns beyond simple physics.

More recent advances in MOT incorporate deep learning to improve association and ID preservation, often by learning detection feature embeddings, motion representations, or similarity metrics that go beyond simple geometric proximity \cite{luo2021multiple} \cite{teye2025motdetr}. Recent research based on methods such as self-supervised and unsupervised learning, have attempted to learn data association without ID annotations \cite{shuai2022idfree}\cite{huang2023supervised}\cite{meng2023unsupervised}. Yet these are in the emerging phase and therefore comprise fewer well-established benchmarks. Further, direct extension of deep learning-based tracking from image to point cloud domain remains non-trivial due to the irregular and sparse nature of point cloud data. 

\subsection{Research gap}\label{subsec:sotagap}

Although previous works have made significant progress in LiDAR-based 3D object detection and multi-object tracking, most published works benchmark performance on autonomous driving applications with extensive evaluations for generic object categories. There remains a notable gap when it comes to detecting and tracking human targets from overhead LiDAR viewpoints, and the unique challenges pertaining to person detection in different environments (particularly outside the domain of autonomous driving scenarios) are not thoroughly addressed by existing implementations. Person detection using LiDAR has been predominantly studied for vehicle or wheeled robot-mounted scenarios which typically comprise frontal sensor perspectives. In comparison, the overhead view yields contrasting point density distributions which can further deteriorate the performance of detection and data association if not explicitly accounted for in the model design. Subsequently, there is also a lack of open-source datasets annotated for person detection and tracking, specifically with an overhead LiDAR viewpoint in indoor work environments such as the overhead crane scenario addressed in the proposed research.

\section{Methods}\label{sec:methods}

\subsection{Problem setting and system overview}\label{subsec:problem}

\begin{figure}[htbp]
\centering
\includegraphics[width=3.45in]{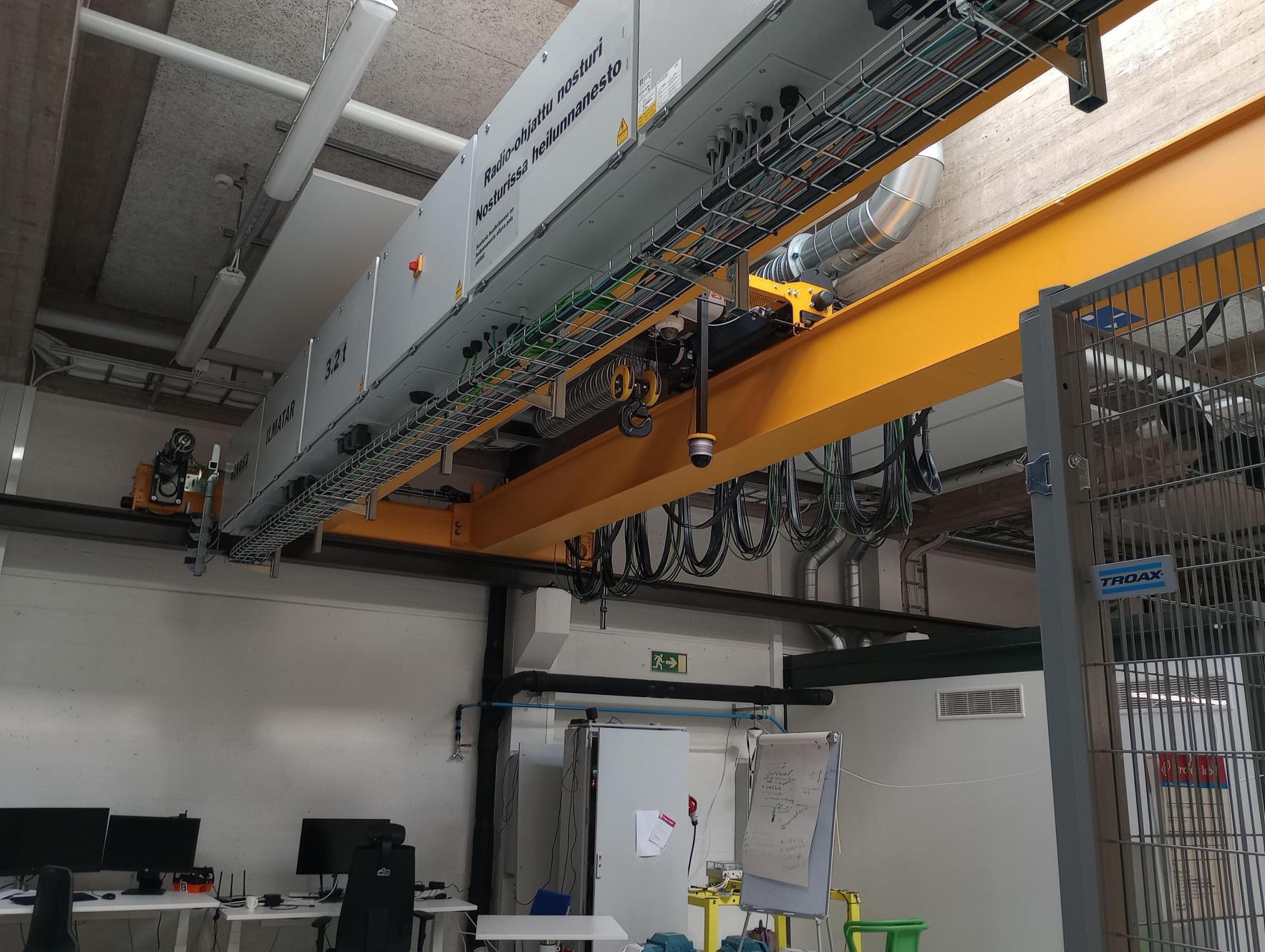}
\caption{Workspace depicting the installed LiDAR on the overhead crane.}
\label{installation_picture}
\end{figure}

This paper addresses the problem of person detection using an overhead-mounted static LiDAR sensor as shown in Figure \ref{installation_picture}. The task is formulated within a prediction framework, where a deep learning–based 3D detector processes sequential point clouds to output per-frame human instances (with an ROI up to $4.5$ m from the sensor) in each frame. A subsequent lightweight tracker associates the detections over time to maintain their IDs and trajectories. 

\begin{figure}[htbp]
\centering
\includegraphics[width=3.45in]{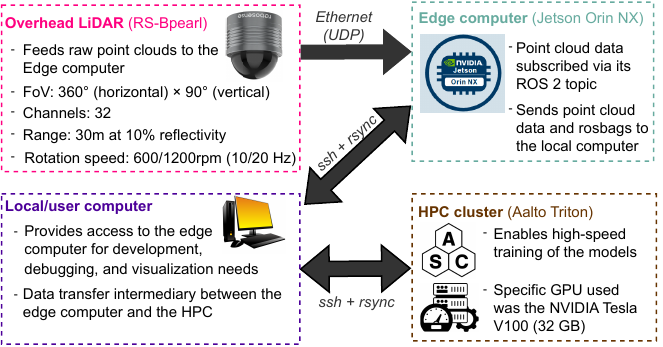}
\caption{Overview of the setup.}
\label{olidar_setup}
\end{figure}

The overhead viewpoint of this indoor work environment constitutes a domain shift from conventional frontal type (e.g. vehicle-centric) LiDAR perception pipelines. Henceforth, the research is motivated on transfer learning and a comparative evaluation of detectors which we adapt for this setting. To facilitate the development of the study, a point cloud dataset was collected from the installed setup shown in Figure \ref{olidar_setup} which consists of the LiDAR connected to an edge unit. The LiDAR was mounted at a height of $2.94$m from the ground. Within the scope of this study, the edge computer primarily enables the extraction of captured data for offline training and evaluation.

\subsection{Model architecture}\label{subsec:model}

\begin{figure}[!ht]
\centering
\includegraphics[width=3.45in]{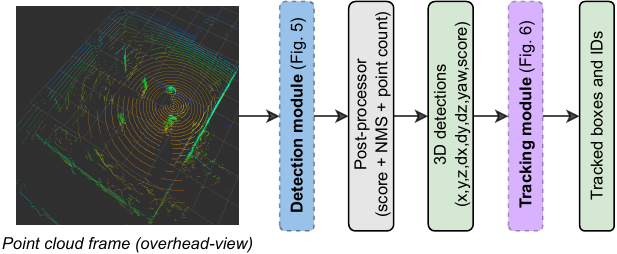}
\caption{Proposed model architecture of the detection and tracking framework.}
\label{model_arch_overall}
\end{figure}

The high-level blocks of the proposed model architecture are illustrated in Figure \ref{model_arch_overall}. The LiDAR-based detector outputs per-frame detections of human targets and the subsequent tracking module associates the detections across consecutive time steps to generate person trajectories and maintain their IDs. 

\subsubsection{Detection module}\label{subsec:det_model}

\begin{figure}[htbp]
\centering
\includegraphics[width=2.5in]{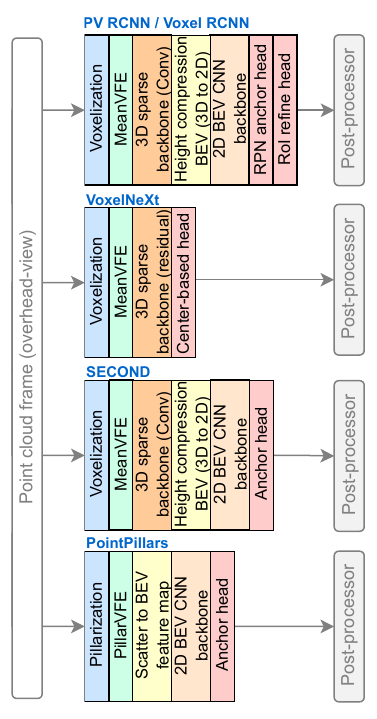}
\caption{Architectures of the candidate detectors. Each model is run independently on the same input point clouds.}
\label{detection_architecture}
\end{figure}

For person detection, we adapt and evaluate multiple LiDAR-based 3D detection modules as shown in Figure \ref{detection_architecture}. Our overhead-view configuration differs from the vehicle-centric viewpoint of common large-scale autonomous driving datasets, in which these detectors have been trained and evaluated by their respective authors. Thus, for a fair evaluation, we leverage transfer learning by fine-tuning each detector from a pretrained checkpoint on our site-specific overhead view dataset.

The primary criteria for selecting the candidate detection architectures are:
{
\renewcommand{\labelenumi}{\roman{enumi}.}
\begin{enumerate}
    \item compatibility with BEV/voxel representations to capture top-down geometry.
    \item support for low-latency deployment to enable online inference in downstream tasks such as real-time personnel monitoring.
    \item documented open-source implementation to enable efficient domain transfer.
    \item reported evidence on standard 3D detection benchmarks, ideally for human targets as well.
\end{enumerate}
}

Based on the above criteria, we select representative detection architectures which are listed below with a brief description:
\begin{enumerate}
    \item PointPillars (PP) \cite{Lang2019CVPR}: groups points into vertical columns (pillars), places the results onto a BEV feature image, and employs a 2D CNN backbone to output 3D person detections. We consider PP to be a suitable choice as \textbf{(i)} its BEV-based architecture emphasizing the horizontal spatial layout aligns best for our top-down viewpoint, and \textbf{(ii)} it is a widely adopted baseline for person detection.
    \item SECOND (SEC) \cite{yan2018second}: widely adopted voxel baseline, which we employ to first voxelize each point cloud, then use sparse 3D convolutions for efficient feature extraction, and finally yield BEV-based person predictions.
    \item PV-RCNN \cite{Shi_2020_CVPR}: two-stage pipeline which we use to first generate 3D box candidates using voxel-based features, and then refine them using point-based features extracted from within each candidate region. Its architecture is aimed at improving localization and could hence be relevant for humans as they are relatively smaller targets. 
    \item VoxelNeXt \cite{chen2023voxelnext}: fully sparse pipeline which we employ to predict human targets directly from sparse voxel features, reducing reliance on dense intermediate representations and improving efficiency.
    \item Voxel RCNN \cite{deng2021voxelrcnn}: another two-stage pipeline where we first generate 3D box candidates with a voxel backbone from a BEV representation, and then refine each candidate by extracting voxel features from within each candidate region. 

\end{enumerate}
It must also be highlighted that the difference between Voxel RCNN and PV RCNN is in the refinement stage. Voxel RCNN refines each proposal using features sampled directly from the voxel feature maps inside the 3D box. In contrast, PV-RCNN first selects a set of representative 3D points (keypoint set abstraction) and enriches them with information from the voxel network (point–voxel fusion), then uses these point-enhanced features for proposal refinement.

\subsubsection{Tracking module}\label{subsec:modeltracker}

\begin{figure}[!ht]
\centering
\includegraphics[width=3.2in]{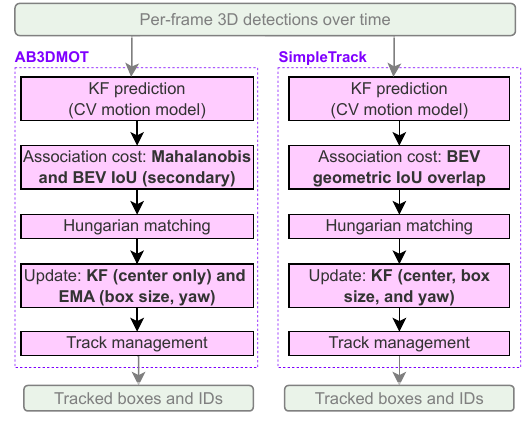}
\caption{Overview of the two tracking-by-detection pipelines, with the inputs being per-frame 3D detections over time.}
\label{tracking_architecture}
\end{figure}

To support downstream monitoring, we adapted two lightweight tracking-by-detection modules: AB3DMOT \cite{weng2020ab3dmot} and SimpleTrack \cite{pang2022simpletrack}. Both trackers take per-frame 3D detections (oriented 3D boxes with confidence scores) as input and maintain a set of tracks over time (Figure \ref{tracking_architecture}). The original source codes of AB3DMOT and SimpleTrack are tailored for KITTI- and Waymo-style dataset loaders respectively. Our implementation preserves the original algorithmic design, with the adaptation focused on processing detections from our overhead-LiDAR data, including file parsing and coordinate/yaw convention handling. 

\begin{figure*}[!ht]
    \centering
    \subfloat[PP]{
        \includegraphics[width=1.31in]{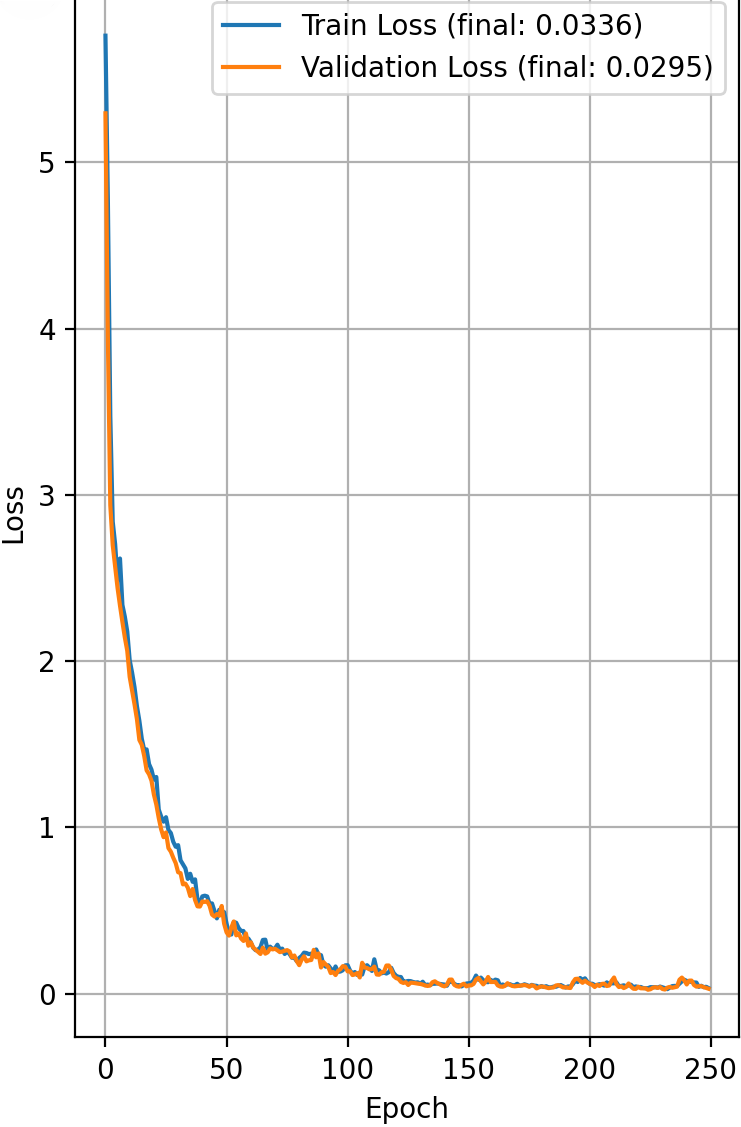}
        \label{loss_curve_pointpillars_150}
    }
    \subfloat[SEC]{
        \includegraphics[width=1.31in]{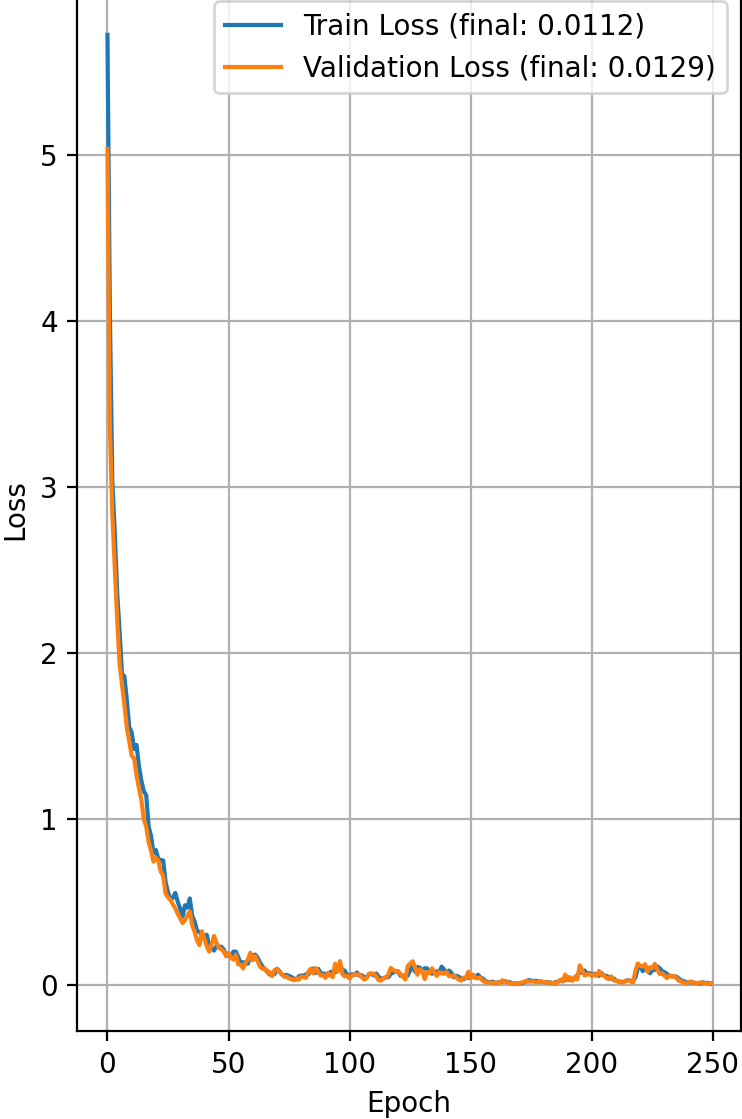}
        \label{loss_curve_second_150}
    }
    \subfloat[PV-RCNN]{
        \includegraphics[width=1.31in]{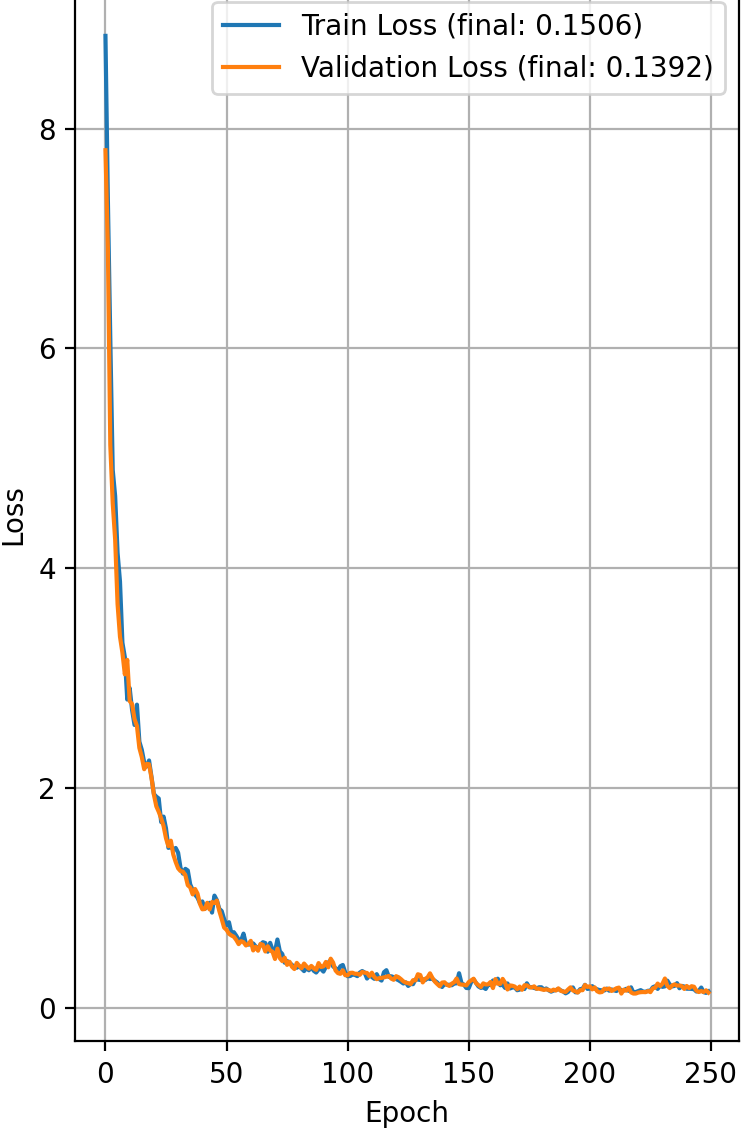}
        \label{loss_curve_pvrcnn_150}
    }
    \subfloat[VoxelNeXt]{
        \includegraphics[width=1.31in]{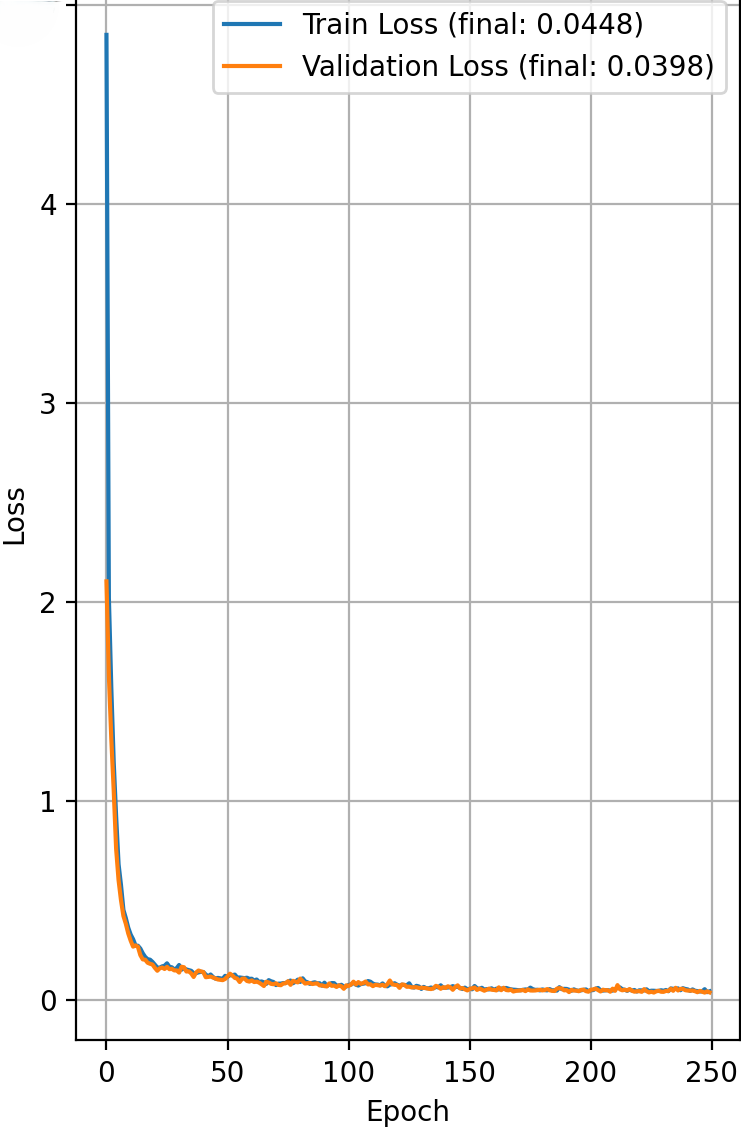}
        \label{loss_curve_voxelnext_150}
    }
    \subfloat[Voxel RCNN]{
        \includegraphics[width=1.31in]{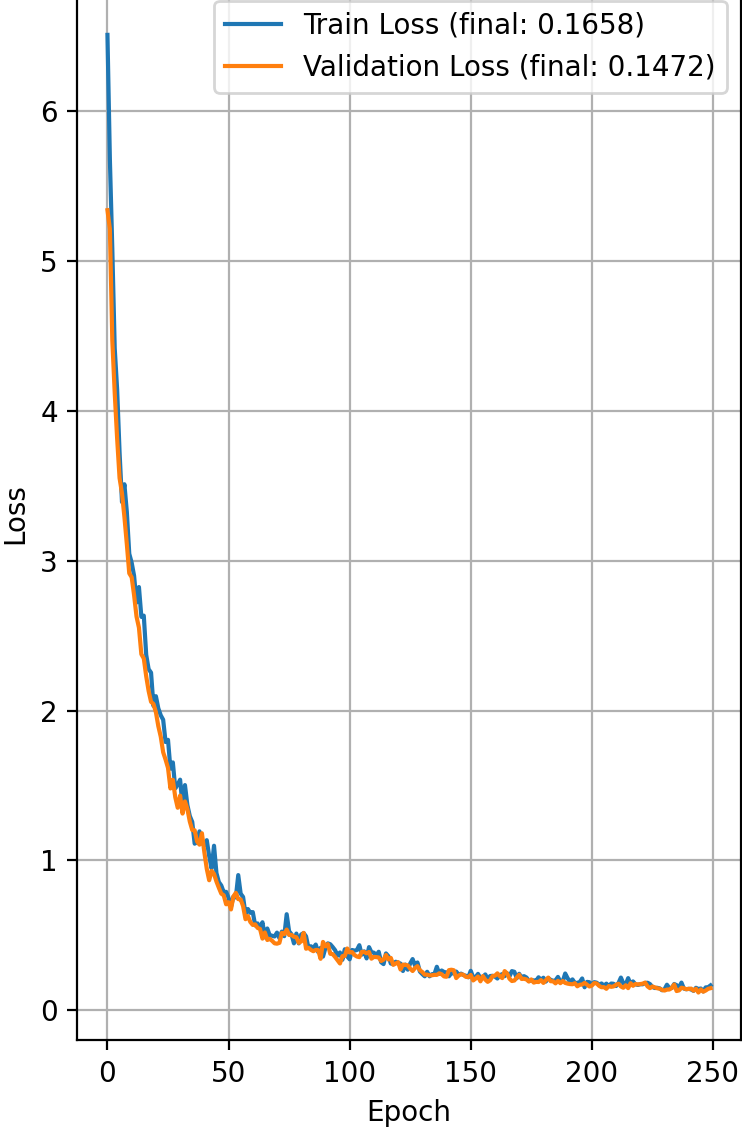}
        \label{loss_curve_voxelrcnn_150}
    }
    \caption{Training and validation plot for each detector based on the architecture denoted in the captions \textbf{(a)}-\textbf{(e)}, obtained by the common data splits of the compiled overhead-view LiDAR dataset.}
    \label{loss_curves}
\end{figure*}

As illustrated in Figure \ref{tracking_architecture}, each track is predicted with a constant-velocity KF using the 3D box center as the measurement. At each time step, predicted tracks are first matched to current detections via Hungarian matching. This is followed by track management, in which unmatched detections create new tentative tracks which are output only after achieving at least $2$ ($min\_hits$) successful matches, and tracks are removed if they remain unmatched for more than $3$ ($max\_age$) frames. This policy allows for short term missed detections while maintaining stable ID assignment over time.

The two trackers differ mainly in how detections are matched to existing tracks and how the estimated bounding boxes are updated over time.
The AB3DMOT-style tracker performs data association based on motion consistency, using the Mahalanobis distance between predicted track positions and detected object centers in 3D space, with BEV-based overlap used as a secondary geometric cue. The box dimensions and yaw are not included in the KF state and instead, are updated at the track level using exponential moving average (EMA). In contrast, SimpleTrack relies directly on geometric overlap between predicted and detected bounding boxes in BEV for association, using fixed overlap thresholds, while retaining the same KF-based motion prediction and track birth/death logic. It follows a more standard KF formulation where the tracked state includes the full 3D box parameters (center, size, and yaw).

Additionally, both trackers are designed for real-time tracking contexts in autonomous driving benchmarks, and that aligns with our motivation for an edge-deployed inference pipeline with the tracking module not becoming a computational bottleneck. Also, both trackers offer a crucial practical benefit of requiring no training data as they are not based on machine learning. Thus, they eliminate labeling overhead and facilitate system transferability to similar overhead LiDAR settings. 
 
\subsection{Training and evaluation}

Training and evaluation in this work are performed separately for the detection and tracking components of the pipeline. Specifically, model training is applied only to the LiDAR-based person detection module, while the tracking module operates without learning and is evaluated independently.

\subsubsection{Detection}
A collected dataset of point cloud frames from the LiDAR setup was manually annotated with 3D bounding boxes for human targets. The annotation was done using labelCloud \cite{sager2022labelcloud}, an open-source tool tailored for labeling point clouds and supporting export formats commonly employed for machine learning pipelines. The training and validation splits comprised 29 and 1 annotated point clouds respectively, with the frames containing only three participants performing arbitrary movements in the monitored area. These two splits were made intentionally small to minimize manual labeling overhead while assessing how effectively the detector candidates can be fine-tuned (transfer learning) to our site-specific overhead-view LiDAR setup.

To summarize the empirical aspects of the final training configuration, each detector was trained for 250 epochs with a batch size of 4 and using the AdamW optimizer. Input point clouds were shuffled during training to improve robustness. Moreover, we initialize the detector backbones from pretrained weights on open-source autonomous driving datasets, such as KITTI \cite{geiger2013vision} and nuScenes \cite{caesar2020nuscenes}, to improve sample efficiency under limited labeled data. The anchors of the detection head, where applicable, were sized as $0.8$m by $0.6$m by $1.73$m, to align it with the typical dimensions of a human under the overhead viewpoint. The final training and validation loss curves are depicted in Figure \ref{loss_curves}.

At test time, the detections were filtered using a confidence threshold of $0.45$, followed by Non-Maximum Suppression to remove duplicate overlapping detections. A prediction was counted as a true positive if it matched a ground-truth box with rotated BEV IoU $\ge 0.10$. To evaluate the accuracy of the detectors, we curated a test set of 76 annotated point clouds, which comprised 10 participants that were not present in the training and validation splits in order to assess the generalization capabilities to new data.

Apart from the configuration adjustments specific to our setup, the remaining settings primarily follow the instructions of OpenPCDet, a toolbox for LiDAR-based perception models \cite{openpcdet2020}. Refer to our repository for full configurations of all adapted detectors.

\subsubsection{Tracking}

To evaluate the tracking pipelines, we used short continuous clips (sequences) extracted from recorded ROS bag data at 3 Hz. We selected multiple clips from different recordings to avoid evaluation only on a single continuous sequence. In total, we manually labeled 80 frames across these clips with 3D bounding boxes for human targets. However, since we did not manually assign IDs, we ran the tracking algorithms on the labeled boxes to automatically assign IDs over time for these clips. These correspond to the pseudo ground truth (pseudo-GT) of our approach which we then use to compare and evaluate the tracking modules. In addition to the tracker implementation details in Section \ref{subsec:modeltracker} and the illustration in Figure \ref{tracking_architecture}, refer to our repository for full configuration details of all adapted tracking algorithms.

\section{Results}

\subsection{Detection}\label{subsec:det_results}
{
\renewcommand{\arraystretch}{1.3}
\begin{table*}[]
\centering
\caption{Evaluation metrics for the adapted detection modules on the test split (true positive if IoU $\ge 0.10$). $\mathbf{r}$ denotes the horizontal radial distance from the LiDAR, and each row reports cumulative performance for human targets within that radius.}
\label{tab:performance_comparison}
\resizebox{\textwidth}{!}{%
\begin{tabular}{l|ccccc|ccccc|ccccc|ccccc|ccccc}
\hline
{\color[HTML]{747474} }                                                                                & \multicolumn{5}{c|}{{\color[HTML]{747474} \textbf{Precision}}}                                                                                                                                                                                                                                                                                                                                                                                                                                 & \multicolumn{5}{c|}{{\color[HTML]{747474} \textbf{Recall}}}                                                                                                                                                                                                                                                                                                                                                                                                                                    & \multicolumn{5}{c|}{{\color[HTML]{747474} \textbf{F1-score}}}                                                                                                                                                                                                                                                                                                                                                                                                                                  & \multicolumn{5}{c|}{{\color[HTML]{747474} \textbf{AP}}}                                                                                                                                                                                                                                                                                                                                                                                                                                        & \multicolumn{5}{c}{{\color[HTML]{747474} \textbf{mIoU}}}                                                                                                                                                                                                                                                                                                                                                                                                                                      \\ \cline{2-26} 
\multirow{-2}{*}{{\color[HTML]{747474} \textbf{\begin{tabular}[c]{@{}l@{}}r \\ (m)\end{tabular}}}} & \multicolumn{1}{l}{{\color[HTML]{747474} \textit{\textbf{PP}}}} & \multicolumn{1}{l}{{\color[HTML]{747474} \textit{\textbf{SEC}}}} & \multicolumn{1}{l}{{\color[HTML]{747474} \textit{\textbf{\begin{tabular}[c]{@{}l@{}}PV-\\ RCNN\end{tabular}}}}} & \multicolumn{1}{l}{{\color[HTML]{747474} \textit{\textbf{\begin{tabular}[c]{@{}l@{}}Voxel\\ Next\end{tabular}}}}} & \multicolumn{1}{l|}{{\color[HTML]{747474} \textit{\textbf{\begin{tabular}[c]{@{}l@{}}Voxel \\ RCNN\end{tabular}}}}} & \multicolumn{1}{l}{{\color[HTML]{747474} \textit{\textbf{PP}}}} & \multicolumn{1}{l}{{\color[HTML]{747474} \textit{\textbf{SEC}}}} & \multicolumn{1}{l}{{\color[HTML]{747474} \textit{\textbf{\begin{tabular}[c]{@{}l@{}}PV-\\ RCNN\end{tabular}}}}} & \multicolumn{1}{l}{{\color[HTML]{747474} \textit{\textbf{\begin{tabular}[c]{@{}l@{}}Voxel\\ Next\end{tabular}}}}} & \multicolumn{1}{l|}{{\color[HTML]{747474} \textit{\textbf{\begin{tabular}[c]{@{}l@{}}Voxel \\ RCNN\end{tabular}}}}} & \multicolumn{1}{l}{{\color[HTML]{747474} \textit{\textbf{PP}}}} & \multicolumn{1}{l}{{\color[HTML]{747474} \textit{\textbf{SEC}}}} & \multicolumn{1}{l}{{\color[HTML]{747474} \textit{\textbf{\begin{tabular}[c]{@{}l@{}}PV-\\ RCNN\end{tabular}}}}} & \multicolumn{1}{l}{{\color[HTML]{747474} \textit{\textbf{\begin{tabular}[c]{@{}l@{}}Voxel\\ Next\end{tabular}}}}} & \multicolumn{1}{l|}{{\color[HTML]{747474} \textit{\textbf{\begin{tabular}[c]{@{}l@{}}Voxel \\ RCNN\end{tabular}}}}} & \multicolumn{1}{l}{{\color[HTML]{747474} \textit{\textbf{PP}}}} & \multicolumn{1}{l}{{\color[HTML]{747474} \textit{\textbf{SEC}}}} & \multicolumn{1}{l}{{\color[HTML]{747474} \textit{\textbf{\begin{tabular}[c]{@{}l@{}}PV-\\ RCNN\end{tabular}}}}} & \multicolumn{1}{l}{{\color[HTML]{747474} \textit{\textbf{\begin{tabular}[c]{@{}l@{}}Voxel\\ Next\end{tabular}}}}} & \multicolumn{1}{l|}{{\color[HTML]{747474} \textit{\textbf{\begin{tabular}[c]{@{}l@{}}Voxel \\ RCNN\end{tabular}}}}} & \multicolumn{1}{l}{{\color[HTML]{747474} \textit{\textbf{PP}}}} & \multicolumn{1}{l}{{\color[HTML]{747474} \textit{\textbf{SEC}}}} & \multicolumn{1}{l}{{\color[HTML]{747474} \textit{\textbf{\begin{tabular}[c]{@{}l@{}}PV-\\ RCNN\end{tabular}}}}} & \multicolumn{1}{l}{{\color[HTML]{747474} \textit{\textbf{\begin{tabular}[c]{@{}l@{}}Voxel\\ Next\end{tabular}}}}} & \multicolumn{1}{l}{{\color[HTML]{747474} \textit{\textbf{\begin{tabular}[c]{@{}l@{}}Voxel \\ RCNN\end{tabular}}}}} \\ \hline
{\color[HTML]{747474} \textbf{1.0}}                                                                    & 1.000                                                           & 1.000                                                            & 1.000                                                                                                           & 1.000                                                                                                             & 1.000                                                                                                               & 0.769                                                           & 0.828                                                            & 0.448                                                                                                           & \textbf{0.966}                                                                                                    & 0.069                                                                                                               & 0.870                                                           & 0.906                                                            & 0.619                                                                                                           & \textbf{0.982}                                                                                                    & 0.129                                                                                                               & 0.713                                                           & 0.823                                                            & 0.448                                                                                                           & \textbf{0.966}                                                                                                    & 0.069                                                                                                               & 0.477                                                           & \textbf{0.586}                                                   & 0.544                                                                                                           & 0.550                                                                                                             & 0.530                                                                                                              \\ \hline
{\color[HTML]{747474} \textbf{2.0}}                                                                    & 0.974                                                           & \textbf{1.000}                                                   & \textbf{1.000}                                                                                                  & 0.983                                                                                                             & 0.500                                                                                                               & 0.603                                                           & 0.887                                                            & 0.403                                                                                                           & \textbf{0.983}                                                                                                    & 0.046                                                                                                               & 0.745                                                           & 0.940                                                            & 0.575                                                                                                           & \textbf{0.983}                                                                                                    & 0.085                                                                                                               & 0.573                                                           & 0.846                                                            & 0.385                                                                                                           & \textbf{0.892}                                                                                                    & 0.023                                                                                                               & 0.427                                                           & 0.557                                                            & 0.550                                                                                                           & \textbf{0.560}                                                                                                    & 0.543                                                                                                              \\ \hline
3.0                                                                                                    & 0.974                                                           & \textbf{1.000}                                                   & \textbf{1.000}                                                                                                  & 0.896                                                                                                             & 0.643                                                                                                               & 0.574                                                           & \textbf{0.882}                                                   & 0.290                                                                                                           & 0.868                                                                                                             & 0.069                                                                                                               & 0.722                                                           & \textbf{0.937}                                                   & 0.450                                                                                                           & 0.882                                                                                                             & 0.125                                                                                                               & 0.547                                                           & \textbf{0.833}                                                   & 0.285                                                                                                           & 0.827                                                                                                             & 0.054                                                                                                               & 0.452                                                           & 0.554                                                            & 0.538                                                                                                           & 0.545                                                                                                             & \textbf{0.556}                                                                                                     \\ \hline
4.0                                                                                                    & 0.967                                                           & 0.994                                                            & \textbf{1.000}                                                                                                  & 0.907                                                                                                             & 0.762                                                                                                               & 0.442                                                           & \textbf{0.848}                                                   & 0.242                                                                                                           & 0.737                                                                                                             & 0.081                                                                                                               & 0.606                                                           & \textbf{0.915}                                                   & 0.390                                                                                                           & 0.813                                                                                                             & 0.146                                                                                                               & 0.452                                                           & \textbf{0.843}                                                   & 0.242                                                                                                           & 0.728                                                                                                             & 0.070                                                                                                               & 0.445                                                           & 0.535                                                            & \textbf{0.536}                                                                                                  & 0.519                                                                                                             & 0.480                                                                                                              \\ \hline
5.0                                                                                                    & 0.967                                                           & 0.935                                                            & \textbf{1.000}                                                                                                  & 0.796                                                                                                             & 0.762                                                                                                               & 0.433                                                           & \textbf{0.847}                                                   & 0.241                                                                                                           & 0.729                                                                                                             & 0.079                                                                                                               & 0.599                                                           & \textbf{0.889}                                                   & 0.389                                                                                                           & 0.761                                                                                                             & 0.143                                                                                                               & 0.451                                                           & \textbf{0.837}                                                   & 0.241                                                                                                           & 0.706                                                                                                             & 0.068                                                                                                               & 0.442                                                           & 0.535                                                            & \textbf{0.536}                                                                                                  & 0.517                                                                                                             & 0.480                                                                                                              \\ \hline
\end{tabular}%
}
\end{table*}
}

To evaluate the performance of the proposed detection modules, we leverage the metrics outlined in Table \ref{tab:performance_comparison}. Note that the recall corresponds to the share of actual persons that are successfully detected. Missed detections dominate risk in our context. AP summarizes the ranking quality of detector confidences over operating points, and F1-score provides a single operating-point balance between false positives and false negatives. The mIoU reflects BEV localization quality for matched detections. Moreover, we leverage distance-based slicing to depict how performance changes across the working radius with our single overhead LiDAR configuration. $\mathbf{r}$ denotes the cumulative horizontal radial range in the LiDAR coordinate frame. For example, at $3.0$ m, we evaluate all detections with $\mathbf{r} \le 3.0$ m Additionally, Table \ref{tab:inference_speed} depicts that all adapted models except PV-RCNN, demonstrate comparable latency for real-time application feasibility. 

Henceforth, primarily dictated by the performance metrics in Table \ref{tab:performance_comparison}, we deduce the VoxelNeXT- and SECOND-based detection modules to be the top performing candidates, with the latter being the most robust for human targets at larger radial distances from the LiDAR. The comparative out-performance of these models is evident primarily from the bolded values for the Recall, F1, and average precision (AP) metrics. The mIoU is largely consistent across distance for all models, while PP is slightly outlying with a comparatively lower mIoU range ($0.43$ - $0.47$).

Moreover, it must be highlighted that we also assessed each detector in a pretrained-only setting by running it with the officially released checkpoint (without any fine-tuning) on our overhead-view LiDAR test split, and evaluating only the Person/Pedestrian class. PointPillars and PV-RCNN produced no true positives under our evaluation protocol, while the available Voxel R-CNN checkpoint was trained only for the Car class. Quantitative results for SECOND and VoxelNeXt are reported in Table []. Despite yielding decent precision values at low IoU thresholds (set to $0.1$ as documented in the original configuration), their overall detection quality remains significantly low, as indicated by the significantly low recall, F1, and AP values in comparison to our adapted models fine-tuned on the overhead-view LiDAR data (Table \ref{tab:performance_comparison}). Thus, the experiment clearly highlighted the domain gap between standard driving datasets and overhead sensing.

{
\renewcommand{\arraystretch}{1.1}
\begin{table}[]
\centering
\caption{Detection performance of the pretrained models (no fine-tuning) on the overhead-view LiDAR test split.}
\label{tab:pretrained_performance}
\resizebox{\columnwidth}{!}{%
\begin{tabular}{l|cc|cc|cc|cc|cc}
\hline
{\color[HTML]{747474} }                                                                                & \multicolumn{2}{c|}{{\color[HTML]{747474} \textbf{Precision}}}                                                                                                                        & \multicolumn{2}{c|}{{\color[HTML]{747474} \textbf{Recall}}}                                                                                                                           & \multicolumn{2}{c|}{{\color[HTML]{747474} \textbf{F1-score}}}                                                                                                                         & \multicolumn{2}{c|}{{\color[HTML]{747474} \textbf{AP}}}                                                                                                                               & \multicolumn{2}{c}{{\color[HTML]{747474} \textbf{mIoU}}}                                                                                                                             \\ \cline{2-11} 
\multirow{-2}{*}{{\color[HTML]{747474} \textbf{\begin{tabular}[c]{@{}l@{}}Dist. \\ (m)\end{tabular}}}} & \multicolumn{1}{l}{{\color[HTML]{747474} \textit{\textbf{\begin{tabular}[c]{@{}l@{}}Voxel\\ NeXt\end{tabular}}}}} & \multicolumn{1}{l|}{{\color[HTML]{747474} \textit{\textbf{SEC}}}} & \multicolumn{1}{l}{{\color[HTML]{747474} \textit{\textbf{\begin{tabular}[c]{@{}l@{}}Voxel\\ NeXt\end{tabular}}}}} & \multicolumn{1}{l|}{{\color[HTML]{747474} \textit{\textbf{SEC}}}} & \multicolumn{1}{l}{{\color[HTML]{747474} \textit{\textbf{\begin{tabular}[c]{@{}l@{}}Voxel\\ NeXt\end{tabular}}}}} & \multicolumn{1}{l|}{{\color[HTML]{747474} \textit{\textbf{SEC}}}} & \multicolumn{1}{l}{{\color[HTML]{747474} \textit{\textbf{\begin{tabular}[c]{@{}l@{}}Voxel\\ NeXt\end{tabular}}}}} & \multicolumn{1}{l|}{{\color[HTML]{747474} \textit{\textbf{SEC}}}} & \multicolumn{1}{l}{{\color[HTML]{747474} \textit{\textbf{\begin{tabular}[c]{@{}l@{}}Voxel\\ NeXt\end{tabular}}}}} & \multicolumn{1}{l}{{\color[HTML]{747474} \textit{\textbf{SEC}}}} \\ \hline
{\color[HTML]{747474} \textbf{1.0}}                                                                    & \textbf{1.000}                                                                                                    & 0.889                                                             & \textbf{0.483}                                                                                                    & 0.286                                                             & \textbf{0.651}                                                                                                    & 0.432                                                             & \textbf{0.435}                                                                                                    & 0.272                                                             & 0.497                                                                                                             & \textbf{0.508}                                                   \\ \hline
{\color[HTML]{747474} \textbf{2.0}}                                                                    & 0.950                                                                                                             & \textbf{0.968}                                                    & \textbf{0.594}                                                                                                    & 0.462                                                             & \textbf{0.731}                                                                                                    & 0.625                                                             & \textbf{0.546}                                                                                                    & 0.451                                                             & 0.522                                                                                                             & \textbf{0.525}                                                   \\ \hline
3.0                                                                                                    & 0.944                                                                                                             & \textbf{0.974}                                                    & 0.515                                                                                                             & \textbf{0.573}                                                    & 0.667                                                                                                             & \textbf{0.721}                                                    & 0.501                                                                                                             & \textbf{0.557}                                                    & \textbf{0.534}                                                                                                    & 0.521                                                            \\ \hline
4.0                                                                                                    & \textbf{0.808}                                                                                                    & 0.694                                                             & 0.426                                                                                                             & \textbf{0.515}                                                    & 0.558                                                                                                             & \textbf{0.591}                                                    & 0.393                                                                                                             & \textbf{0.446}                                                    & 0.513                                                                                                             & \textbf{0.516}                                                   \\ \hline
5.0                                                                                                    & 0.635                                                                                                             & \textbf{0.688}                                                    & 0.429                                                                                                             & \textbf{0.522}                                                    & 0.512                                                                                                             & \textbf{0.594}                                                    & 0.367                                                                                                             & \textbf{0.471}                                                    & 0.505                                                                                                             & \textbf{0.519}                                                   \\ \hline
\end{tabular}%
}
\end{table}
}
\subsection{Tracking}\label{subsec:tracks_results}

The tracking performance was quantified using \textbf{(i)} Multi-Object Tracking Accuracy (MOTA) to summarize overall tracking quality via false positives, false negatives, and ID switches, \textbf{(ii)} Multi-Object Tracking Precision (MOTP) to measure localization accuracy of matched tracks (reported as mean oriented BEV IoU), and \textbf{(iii)} Identification F1 score (IDF1) to assess ID consistency over time by balancing identification precision and recall.

The evaluated metrics are outlined in Table \ref{tab:track_metrics}. Note here that we employed the top 3 detectors for comparison (based on their results outlined in section \ref{subsec:det_results}. Matching uses oriented BEV IoU, and MOTA/IDF1/MOTP are reported at IoU thresholds of $0.3$ and $0.1$. Between the two implemented tracking methods, the AB3DMOT-style tracker demonstrated significantly faster execution, as indicated by the median percentile (p50) of $1.08$ms which is nearly 6 times faster than the SimpleTrack based tracker. Yet, both trackers are well within the latency required for real-time deployment.    

{
\renewcommand{\arraystretch}{1.1}
\begin{table}[]
\centering
\caption{Tracking performance for the modules adapted from AB3DMOT and SimpleTrack on the temporal sequences.}
\label{tab:track_metrics}
\resizebox{\columnwidth}{!}{%
\begin{tabular}{l|l|ccc|ccc}
\hline
{\color[HTML]{747474} }                                       & {\color[HTML]{747474} }                                             & \multicolumn{3}{c|}{{\color[HTML]{747474} \textbf{IoU = 0.3}}}                                                                                                                  & \multicolumn{3}{c}{{\color[HTML]{747474} \textbf{IoU = 0.1}}}                                                                                                                  \\ \cline{3-8} 
\multirow{-2}{*}{{\color[HTML]{747474} \textbf{Tracker}}}     & \multirow{-2}{*}{{\color[HTML]{747474} \textit{\textbf{Detector}}}} & \multicolumn{1}{l}{{\color[HTML]{747474} \textbf{MOTA}}} & \multicolumn{1}{l}{{\color[HTML]{747474} \textbf{IDF1}}} & \multicolumn{1}{l|}{{\color[HTML]{747474} \textbf{MOTP}}} & \multicolumn{1}{l}{{\color[HTML]{747474} \textbf{MOTA}}} & \multicolumn{1}{l}{{\color[HTML]{747474} \textbf{IDF1}}} & \multicolumn{1}{l}{{\color[HTML]{747474} \textbf{MOTP}}} \\ \hline
{\color[HTML]{747474} }                                       & {\color[HTML]{747474} \textit{\textbf{PP}}}                         & 0.12                                                     & 0.53                                                     & 0.46                                                      & 0.38                                                     & 0.68                                                     & 0.40                                                     \\ \cline{2-8} 
{\color[HTML]{747474} }                                       & {\color[HTML]{747474} \textit{\textbf{SEC}}}                        & 0.33                                                     & 0.72                                                     & \textbf{0.58}                                                      & 0.57                                                     & 0.83                                                     & \textbf{0.54}                                                     \\ \cline{2-8} 
\multirow{-3}{*}{{\color[HTML]{747474} \textbf{AB3DMOT}}}     & {\color[HTML]{747474} \textit{\textbf{VoxelNeXt}}}                 & \textbf{0.70}                                            & \textbf{0.87}                                            & 0.55                                                      & \textbf{0.83}                                            & \textbf{0.93}                                            & 0.53                                                     \\ \hline
{\color[HTML]{747474} }                                       & {\color[HTML]{747474} \textit{\textbf{PP}}}                         & 0.26                                                     & 0.58                                                     & 0.46                                                      & 0.52                                                     & 0.74                                                     & 0.41                                                     \\ \cline{2-8} 
{\color[HTML]{747474} }                                       & {\color[HTML]{747474} \textit{\textbf{SEC}}}                        & 0.41                                                     & 0.74                                                     & \textbf{0.57}                                                      & 0.60                                                     & 0.83                                                     & 0.54                                                     \\ \cline{2-8} 
\multirow{-3}{*}{{\color[HTML]{747474} \textbf{SimpleTrack}}} & {\color[HTML]{747474} \textit{\textbf{VoxelNeXt}}}                 & \textbf{0.71}                                            & \textbf{0.86}                                            & \textbf{0.57}                                                      & \textbf{0.81}                                            & \textbf{0.91}                                            & \textbf{0.55}                                                     \\ \hline
\end{tabular}%
}
\end{table}
}

{
\renewcommand{\arraystretch}{1.1}
\begin{table}[]
\centering
\caption{Comparison of the obtained inference speeds (via percentiles) for the adapted modules, executed on a CPU (Intel Xeon Gold 6148 @ 2.40 GHz)}
\label{tab:inference_speed}
\resizebox{\columnwidth}{!}{%
\begin{tabular}{l|ccccc|cc}
\hline
                                         & \multicolumn{5}{c|}{{\color[HTML]{747474} \textbf{Detection}}}                                                                                                                                                                                                                                                                                                                                                                                                                                 & \multicolumn{2}{c}{{\color[HTML]{747474} \textbf{Tracking}}}                                                                                    \\ \cline{2-8} 
\multirow{-2}{*}{}                       & \multicolumn{1}{l}{{\color[HTML]{747474} \textit{\textbf{PP}}}} & \multicolumn{1}{l}{{\color[HTML]{747474} \textit{\textbf{SEC}}}} & \multicolumn{1}{l}{{\color[HTML]{747474} \textit{\textbf{\begin{tabular}[c]{@{}l@{}}PV-\\ RCNN\end{tabular}}}}} & \multicolumn{1}{l}{{\color[HTML]{747474} \textit{\textbf{\begin{tabular}[c]{@{}l@{}}Voxel\\ NeXt\end{tabular}}}}} & \multicolumn{1}{l|}{{\color[HTML]{747474} \textit{\textbf{\begin{tabular}[c]{@{}l@{}}Voxel \\ RCNN\end{tabular}}}}} & \multicolumn{1}{l}{{\color[HTML]{747474} \textit{\textbf{AB3DMOT}}}} & \multicolumn{1}{l}{{\color[HTML]{747474} \textit{\textbf{SimpleTrack}}}} \\ \hline
{\color[HTML]{747474} \textbf{p50 (ms)}} & 34                                                              & 46                                                               & 229                                                                                                            & 35                                                                                                                & \textbf{32}                                                                                                                   & \textbf{1.08}                                                                 & 6.30                                                                     \\ \hline
{\color[HTML]{747474} \textbf{p90 (ms)}} & 37                                                              & 47                                                               & 232                                                                                                            & \textbf{36}                                                                                                                & 38                                                                                                                   & \textbf{1.26}                                                                 & 6.66                                                                     \\ \hline
\end{tabular}%
}
\end{table}
}
\section{Discussion}

The distance-sliced detection results (Table \ref{tab:performance_comparison}) across our working radius (up to $4.5$m) suggest that modern voxel-based backbones (VoxelNeXt and SECOND), compared to the remaining models, generalize better to the sparsity and perspective induced by a single overhead LIDAR. A key trend is that VoxelNeXt is best for $<3$ m range as observed by its respective recall, F1 and AP values. Although it remains competitive beyond this range, SECOND demonstrates more robustness at larger radial distances ($>3$ m) where point density decreases with range and, due to the overhead viewpoint, this can lead to partial surface coverage of the human body. This distinction matters for deployment: if the intended application prioritizes high detection confidence of persons in the near field, VoxelNeXt would be the most suitable choice. Alternatively, if the application prioritizes stable coverage across the workspace limits, SECOND would be better as it provides more reliable performance even at longer distances. These outcomes dictate the potential operating envelop in practice for our single overhead-view LiDAR configuration.

Moving to the tracking results, we observe that the tracking quality is broadly similar between the two implemented modules, with the dominant factor being the upstream detector quality. Thus, the best results were obtained with the adapted VoxelNeXt detector outputs, highlighting that for application contexts similar to ours, improving detection quality is critical to improve end-to-end tracking (Table \ref{tab:track_metrics}). Further, analyzing the results for two IoU thresholds was useful as it revealed how sensitive the pipeline was to localization strictness. Specifically, lowering the IoU threshold from $0.3$ to $0.1$ increased MOTA and IDF1 for all detector-tracker combinations which indicates that a portion of errors at IoU = $0.3$ are attributable to modest BEV misalignment rather than complete failures in association. On the other hand, it increases the risk of wrong person associations (e.g. ID swaps) in close interactions.

Additionally, despite our inference benchmarking (Table \ref{tab:inference_speed}) being collected on the NVIDIA V100 rather than the target edge device shown in Figure \ref{olidar_setup}, it remains valuable as a consistent, hardware-controlled comparison of relative model latency under identical conditions. Thus, it provides a practical indicator to estimate which architectures remain feasible for real-time applicability. However, benchmarking on the edge device is a crucial next step as it would provide a more accurate estimate of end-to-end real-world runtime feasibility. Furthermore, our reporting of latency with percentiles is informative as they are robust to system-level latency effects such as I/O and caching.

A limitation with the current study is that our dataset was quantity-limited. Expanding the dataset would be a recommended future action in transforming the research from a site-specific benchmark to more broader operating conditions. The horizontal radial distance for evaluation (Table \ref{tab:performance_comparison}) was also capped at $4.5$ m as it was a practical constraint of our experimental workspace (thus, human targets observed only up to this range). Future work could push the distance boundaries to identify the point at which LiDAR point sparsity becomes too low to sufficiently represent human targets for detection. Furthermore, the evaluation approach for the tracking modules could be further improved for our application context, by converting the pseudo-GT sequences into true tracking ground truth by manually validating and correcting ID continuity for all frames. This would enable more tuning of the trackers and consequently strengthen the benchmark comparisons.

Overall, this study indicates that reliable overhead-view LiDAR person detection and tracking can be accurately implemented in an indoor workspace. The best-performing detector configurations achieved AP values up to $0.84$ when evaluating human targets within our full operating radius of $5.0$m from the LiDAR (horizontally). With shorter monitoring radii of $1.0$m and $2.0$m, the performance was stronger as indicated by top AP values of $0.97$ and $0.89$ respectively. The end-to-end tracking results further showed that tracking-by-detection methods can maintain the track IDs of persons reasonably well over time, and that overall tracking performance is mainly driven by the quality of the detections. At the same time, precise localization of people proved difficult, as indicated by the lower MOTP-values. This was likely due to the commonly faced difficulty of accurately placing 3D bounding boxes on people in LiDAR point clouds. IoU-based matching for human targets can change noticeably even with small position or orientation errors in BEV. To facilitate further development, we release our dataset and code as open-source. Future work should focus on more extensive datasets for overhead person detection from LiDAR point clouds, as well as experiments in more diverse and dynamic industrial environments.

\section*{Acknowledgment}

We acknowledge the funding provided by the Techboost and TwinFlow projects, the computational resources provided by Aalto University’s Scientific Computing Group, and the technical expertise provided by S.M. Hari Prasanth.

\bibliographystyle{IEEEtran}
\balance
\bibliography{refs}

@ARTICLE{Fang2022occlusion,
  author={Li, Fang and Li, Xueyuan and Liu, Qi and Li, Zirui},
  journal={IEEE Access}, 
  title={Occlusion Handling and Multi-Scale Pedestrian Detection Based on Deep Learning: A Review}, 
  year={2022},
  volume={10},
  number={},
  pages={19937-19957},
  doi={10.1109/ACCESS.2022.3150988}}

@INPROCEEDINGS{Linder2021cross,
  author={Linder, Timm and Vaskevicius, Narunas and Schirmer, Robert and Arras, Kai O.},
  booktitle={2021 IEEE/RSJ International Conference on Intelligent Robots and Systems (IROS)}, 
  title={Cross-Modal Analysis of Human Detection for Robotics: An Industrial Case Study}, 
  year={2021},
  volume={},
  number={},
  pages={971-978},
  doi={10.1109/IROS51168.2021.9636158}}

@article{ahmed2021top,
  title={Top view multiple people tracking by detection using deep SORT and YOLOv3 with transfer learning: within 5G infrastructure},
  author={Ahmed, Imran and Ahmad, Misbah and Ahmad, Awais and Jeon, Gwanggil},
  journal={International Journal of Machine Learning and Cybernetics},
  volume={12},
  number={11},
  pages={3053--3067},
  year={2021},
  publisher={Springer}
}

@InProceedings{Lang2019CVPR,
    author = {Lang, Alex H. and Vora, Sourabh and Caesar, Holger and Zhou, Lubing and Yang, Jiong and Beijbom, Oscar},
    title = {PointPillars: Fast Encoders for Object Detection From Point Clouds},
    booktitle = {Proceedings of the IEEE/CVF Conference on Computer Vision and Pattern Recognition (CVPR)},
    month = {June},
    year = {2019}
    }

@inproceedings{weng2020ab3dmot,
  title={3d multi-object tracking: A baseline and new evaluation metrics},
  author={Weng, Xinshuo and Wang, Jianren and Held, David and Kitani, Kris},
  booktitle={2020 IEEE/RSJ International Conference on Intelligent Robots and Systems (IROS)},
  pages={10359--10366},
  year={2020},
  organization={IEEE}
}

@INPROCEEDINGS{zhou2019fvnet,
  author={Zhou, Jie and Tan, Xin and Shao, Zhiwen and Ma, Lizhuang},
  booktitle={2019 12th International Congress on Image and Signal Processing, BioMedical Engineering and Informatics (CISP-BMEI)}, 
  title={FVNet: 3D Front-View Proposal Generation for Real-Time Object Detection from Point Clouds}, 
  year={2019},
  volume={},
  number={},
  pages={1-8},
  doi={10.1109/CISP-BMEI48845.2019.8965844}}

@InProceedings{meyer2019lasernet,
author = {Meyer, Gregory P. and Laddha, Ankit and Kee, Eric and Vallespi-Gonzalez, Carlos and Wellington, Carl K.},
title = {LaserNet: An Efficient Probabilistic 3D Object Detector for Autonomous Driving},
booktitle = {Proceedings of the IEEE/CVF Conference on Computer Vision and Pattern Recognition (CVPR)},
month = {June},
year = {2019}
}

@InProceedings{zhou2018voxelnet,
author = {Zhou, Yin and Tuzel, Oncel},
title = {VoxelNet: End-to-End Learning for Point Cloud Based 3D Object Detection},
booktitle = {Proceedings of the IEEE Conference on Computer Vision and Pattern Recognition (CVPR)},
month = {June},
year = {2018}
}

@article{yan2018second,
  title={Second: Sparsely embedded convolutional detection},
  author={Yan, Yan and Mao, Yuxing and Li, Bo},
  journal={Sensors},
  volume={18},
  number={10},
  pages={3337},
  year={2018},
  publisher={Multidisciplinary Digital Publishing Institute}
}

@inproceedings{deng2021voxelrcnn,
  title={Voxel r-cnn: Towards high performance voxel-based 3d object detection},
  author={Deng, Jiajun and Shi, Shaoshuai and Li, Peiwei and Zhou, Wengang and Zhang, Yanyong and Li, Houqiang},
  booktitle={Proceedings of the AAAI conference on artificial intelligence},
  volume={35},
  number={2},
  pages={1201--1209},
  year={2021}
}

@inproceedings{zheng2021cia,
  title={Cia-ssd: Confident iou-aware single-stage object detector from point cloud},
  author={Zheng, Wu and Tang, Weiliang and Chen, Sijin and Jiang, Li and Fu, Chi-Wing},
  booktitle={Proceedings of the AAAI conference on artificial intelligence},
  volume={35},
  number={4},
  pages={3555--3562},
  year={2021}
}

@InProceedings{chen2023voxelnext,
    author    = {Chen, Yukang and Liu, Jianhui and Zhang, Xiangyu and Qi, Xiaojuan and Jia, Jiaya},
    title     = {VoxelNeXt: Fully Sparse VoxelNet for 3D Object Detection and Tracking},
    booktitle = {Proceedings of the IEEE/CVF Conference on Computer Vision and Pattern Recognition (CVPR)},
    month     = {June},
    year      = {2023},
    pages     = {21674-21683}
}

@InProceedings{shi2019pointrcnn,
    author = {Shi, Shaoshuai and Wang, Xiaogang and Li, Hongsheng},
    title = {PointRCNN: 3D Object Proposal Generation and Detection From Point Cloud},
    booktitle = {Proceedings of the IEEE/CVF Conference on Computer Vision and Pattern Recognition (CVPR)},
    month = {June},
    year = {2019}
    }

@InProceedings{pan2021pointformer,
    author    = {Pan, Xuran and Xia, Zhuofan and Song, Shiji and Li, Li Erran and Huang, Gao},
    title     = {3D Object Detection With Pointformer},
    booktitle = {Proceedings of the IEEE/CVF Conference on Computer Vision and Pattern Recognition (CVPR)},
    month     = {June},
    year      = {2021},
    pages     = {7463-7472}
}

@InProceedings{shi2020pointgnn,
    author = {Shi, Weijing and Rajkumar, Raj},
    title = {Point-GNN: Graph Neural Network for 3D Object Detection in a Point Cloud},
    booktitle = {Proceedings of the IEEE/CVF Conference on Computer Vision and Pattern Recognition (CVPR)},
    month = {June},
    year = {2020}
    }

@InProceedings{blanch2024surveillance,
    author    = {Blanch, Miquel Romero and Li, Zenjie and Escalera, Sergio and Nasrollahi, Kamal},
    title     = {LiDAR-Assisted 3D Human Detection for Video Surveillance},
    booktitle = {Proceedings of the IEEE/CVF Winter Conference on Applications of Computer Vision (WACV) Workshops},
    month     = {January},
    year      = {2024},
    pages     = {123-131}
}

@article{wang2022lidar,
  title={LiDAR-based dense pedestrian detection and tracking},
  author={Wang, Wenguang and Chang, Xiyuan and Yang, Jihuang and Xu, Gaofei},
  journal={Applied Sciences},
  volume={12},
  number={4},
  pages={1799},
  year={2022},
  publisher={MDPI}
}

@article{na2023real,
  title={Real-time 3D multi-pedestrian detection and tracking using 3D LiDAR point cloud for mobile robot},
  author={Na, Ki-In and Park, Byungjae},
  journal={ETRI Journal},
  volume={45},
  number={5},
  pages={836--846},
  year={2023},
  publisher={Wiley Online Library}
}

@inproceedings{zhang2022bytetrack,
  title={Bytetrack: Multi-object tracking by associating every detection box},
  author={Zhang, Yifu and Sun, Peize and Jiang, Yi and Yu, Dongdong and Weng, Fucheng and Yuan, Zehuan and Luo, Ping and Liu, Wenyu and Wang, Xinggang},
  booktitle={European conference on computer vision},
  pages={1--21},
  year={2022},
  organization={Springer}
}

@article{luo2021multiple,
  title={Multiple object tracking: A literature review},
  author={Luo, Wenhan and Xing, Junliang and Milan, Anton and Zhang, Xiaoqin and Liu, Wei and Kim, Tae-Kyun},
  journal={Artificial intelligence},
  volume={293},
  pages={103448},
  year={2021},
  publisher={Elsevier}
}

@article{teye2025motdetr,
  title={LiDAR MOT-DETR: A LiDAR-based Two-Stage Transformer for 3D Multiple Object Tracking},
  author={Teye, Martha Teiko and Maoz, Ori and Rottmann, Matthias},
  journal={arXiv preprint arXiv:2505.12753},
  year={2025}
}

@InProceedings{shuai2022idfree,
    author    = {Shuai, Bing and Li, Xinyu and Kundu, Kaustav and Tighe, Joseph},
    title     = {Id-Free Person Similarity Learning},
    booktitle = {Proceedings of the IEEE/CVF Conference on Computer Vision and Pattern Recognition (CVPR)},
    month     = {June},
    year      = {2022},
    pages     = {14689-14699}
}

@InProceedings{huang2023supervised,
    author    = {Huang, Kaer and Lertniphonphan, Kanokphan and Chen, Feng and Li, Jian and Wang, Zhepeng},
    title     = {Multi-Object Tracking by Self-Supervised Learning Appearance Model},
    booktitle = {Proceedings of the IEEE/CVF Conference on Computer Vision and Pattern Recognition (CVPR) Workshops},
    month     = {June},
    year      = {2023},
    pages     = {3163-3169}
}

@InProceedings{meng2023unsupervised,
    author    = {Meng, Sha and Shao, Dian and Guo, Jiacheng and Gao, Shan},
    title     = {Tracking without Label: Unsupervised Multiple Object Tracking via Contrastive Similarity Learning},
    booktitle = {Proceedings of the IEEE/CVF International Conference on Computer Vision (ICCV)},
    month     = {October},
    year      = {2023},
    pages     = {16264-16273}
}

@article{sager2022labelcloud,
    year = 2022,
    month = {mar},
    volume = {19},
    number = {6},
    pages = {1191--1206},
    author = {Christoph Sager and Patrick Zschech and Niklas Kuhl},
    title = {{labelCloud}: A Lightweight Labeling Tool for Domain-Agnostic 3D Object Detection in Point Clouds},
    journal = {Computer-Aided Design and Applications}
}

@InProceedings{Shi_2020_CVPR,
author = {Shi, Shaoshuai and Guo, Chaoxu and Jiang, Li and Wang, Zhe and Shi, Jianping and Wang, Xiaogang and Li, Hongsheng},
title = {PV-RCNN: Point-Voxel Feature Set Abstraction for 3D Object Detection},
booktitle = {Proceedings of the IEEE/CVF Conference on Computer Vision and Pattern Recognition (CVPR)},
month = {June},
year = {2020}
}

@article{geiger2013vision,
  title={Vision meets robotics: The kitti dataset},
  author={Geiger, Andreas and Lenz, Philip and Stiller, Christoph and Urtasun, Raquel},
  journal={The international journal of robotics research},
  volume={32},
  number={11},
  pages={1231--1237},
  year={2013},
  publisher={Sage Publications Sage UK: London, England}
}

@inproceedings{caesar2020nuscenes,
  title={nuscenes: A multimodal dataset for autonomous driving},
  author={Caesar, Holger and Bankiti, Varun and Lang, Alex H and Vora, Sourabh and Liong, Venice Erin and Xu, Qiang and Krishnan, Anush and Pan, Yu and Baldan, Giancarlo and Beijbom, Oscar},
  booktitle={Proceedings of the IEEE/CVF conference on computer vision and pattern recognition},
  pages={11621--11631},
  year={2020}
}

@misc{openpcdet2020,
    title={OpenPCDet: An Open-source Toolbox for 3D Object Detection from Point Clouds},
    author={OpenPCDet Development Team},
    howpublished = {\url{https://github.com/open-mmlab/OpenPCDet}},
    year={2020}
}

@inproceedings{pang2022simpletrack,
  title={Simpletrack: Understanding and rethinking 3d multi-object tracking},
  author={Pang, Ziqi and Li, Zhichao and Wang, Naiyan},
  booktitle={European Conference on Computer Vision},
  pages={680--696},
  year={2022},
  organization={Springer}
}

\end{document}